\documentclass[10pt,twocolumn,letterpaper]{article}

\usepackage[pagenumbers]{cvpr} 

\definecolor{cvprblue}{rgb}{0.21,0.49,0.74}
\usepackage[pagebackref,breaklinks,colorlinks,allcolors=cvprblue]{hyperref}
\usepackage{pifont}

\usepackage{multirow}
\usepackage{multicol}
\usepackage[table]{xcolor}
\def\confName{CVPR}
\def\confYear{2027}
\usepackage[table]{xcolor}  

\title{AIFR: Adaptive Identity Feature Recovery for Diffusion-Based Makeup and Anime Portraits}

\author{
Linyan Dai$^{1}$,\quad
Xinwei Zhang$^{1}$,\quad
Yaxin Xiao$^{1}$,\quad
Mengyu Sun$^{1}$,\\[-1pt]
Xinwei Liu$^{1}$,\quad
Renyang Liu$^{2}$,\quad
Qingqing Ye$^{1}$,\quad
Haibo Hu$^{1,\ast}$\\[-1pt]
{\small
$^{1}$The Hong Kong Polytechnic University
\qquad
$^{2}$National University of Singapore
}
}

\makeatletter

\patchcmd{\@maketitle}
  {\vskip .375in}
  {\vskip .20in}
  {}
  {\PackageWarning{AIFR}{Failed to patch top spacing}}

\patchcmd{\@maketitle}
  {\vspace*{24pt}}
  {\vspace*{4pt}}
  {}
  {\PackageWarning{AIFR}{Failed to patch title spacing}}

\patchcmd{\@maketitle}
  {\vspace*{12pt}}
  {\vspace*{6pt}}
  {}
  {\PackageWarning{AIFR}{Failed to patch author spacing}}

\makeatother
\begin{document}
\maketitle
\begin{abstract}
Diffusion-based makeup and anime portraits are increasingly popular in online content creation (e.g., virtual makeup editing and anime avatar generation), but their altered appearances can hinder face recognition (FR) in identity-based content screening. Existing methods mainly adapt FR encoders or restore input images. The former require costly gallery re-encoding, while the latter may be incomplete or distort identity-related facial details. To address these limitations, we aim to correct identity features within a frozen FR model to recover its output embeddings in its original space. Appearance changes vary in strength and affect facial regions differently. To this end, we propose \textbf{Adaptive Identity Feature Recovery (AIFR)}, a framework that inserts a lightweight recovery module at a single intermediate layer to predict sample-level correction strength and spatially adaptive residual updates. Module training uses fused identity targets to guide embedding recovery in the original FR space and an identity-shift score to regulate correction strength and limit unnecessary changes. At inference, AIFR processes a single portrait in one forward pass without reference images, producing recovered embeddings that can be directly matched with existing gallery templates without re-encoding. With only 1.22M trainable parameters, AIFR improves the average TAR at FAR=$10^{-4}$ from 66.1\% to 90.3\% over the original ArcFace, and further experiments show improvements across unseen makeup styles, stylization strengths, and generators. Beyond recognition, the recovered embeddings can directly condition a pretrained Arc2Face model to generate faces with improved identity preservation, showing compatibility with embedding-conditioned downstream generation.

\end{abstract}    
\section{Introduction}
\label{sec:intro}

\begin{figure}
    \centering
    \includegraphics[width=1\linewidth]{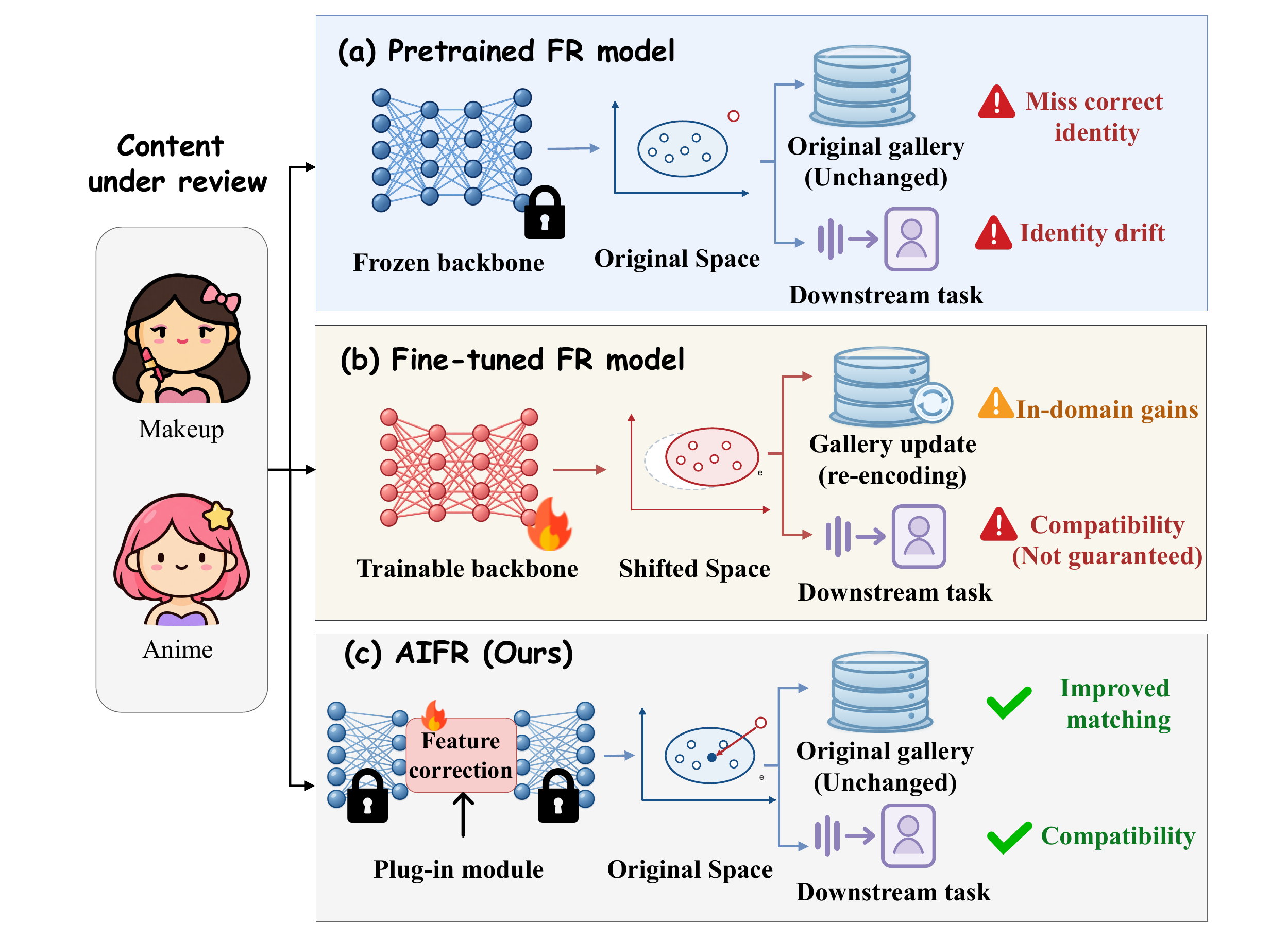}
    \caption{Comparison of three deployment settings for identity matching and downstream compatibility on appearance-altered portraits. (a) Pretrained FR may fail on makeup and anime portraits; (b) Fine-tuning can improve in-domain identity matching with limited generalization, and compatibility with the downstream task is not guaranteed; (c) AIFR improves identity matching across appearance conditions in the original space, supporting compatibility with embedding-conditioned downstream models.}
    \label{application}
\end{figure}
Diffusion models can change a person's facial appearance while preserving the original identity in diverse ways, ranging from realistic makeup transfer to anime-style portrait generation~\cite{zhang2025stablemakeup,ruan2025mad,wang2024instantid,guo2024pulid,gao2026FRAM,zhu2025fluxmakeup,li2024photomaker,sun2024shmt}. Popular applications such as YouCam and Snapchat offer virtual makeup try-on and personalized AI portrait creation for social sharing, respectively~\cite{perfectcorpMakeupTransfer,snap2024aisnaps}. Although such appearance-altered portraits remain recognizable to humans, appearance changes can shift the identity features extracted by face recognition (FR) models, impairing identity matching~\cite{wang2020fm2unet,2026styleid} (Fig.~\ref{application}(a)). This creates a practical challenge for online content screening, where a face under review may need to be matched against an existing gallery of public figures or enrolled creators~\cite{meta2024celebbait,deepface2025}. For example, Meta has reported using FR systems to compare faces in suspected celebrity-bait advertisements with public figures' Facebook and Instagram profile images~\cite{meta2024celebbait}.

Existing approaches aim to improve identity matching for appearance-altered portraits through image
restoration or recognizer adaptation. Image restoration attempts to remove makeup or stylization before recognition~\cite{gavas2026diffclean,ruan2025mad,samc2020}, but incomplete removal and restoration artifacts can
still distort identity cues~\cite{gavas2026diffclean}. 
Alternatively, recognizer adaptation can improve
recognition performance by retraining or fine-tuning FR encoders within the target domain~\cite{
wang2020fm2unet,peng2025stylizedface,2026styleid}. However, as shown in Fig.~\ref{application}(b), gains from recognizer adaptation may not generalize to unseen appearance conditions. Without explicit compatibility constraints, the updated embeddings may also no longer align with the deployed FR space, potentially requiring gallery re-encoding~\cite{shen2020bct,meng2021lce} and adaptation of downstream models trained on that space, such as embedding-conditioned face
generation and personalization ~\cite{papantoniou2024arc2face,facereconstruction2025,peng2024portraitbooth,shiohara2024face2diffusion}.

To address these limitations, our key idea is to directly recover identity features within a pretrained FR model, enabling recognition with existing galleries and compatibility with downstream models. Effective recovery should account for variations in appearance changes across both samples and facial regions. Makeup often affects localized facial attributes, while anime stylization may introduce broader changes in facial texture and structure~\cite{gao2026FRAM,peng2025stylizedface,2026styleid}. These variations motivate adaptive correction that adjusts the overall recovery strength for each input and the feature updates at individual spatial locations. Intermediate convolutional features retain spatially organized facial cues~\cite{phan2022deepface,qiu2021end2end}, enabling such selective corrections before the final identity embedding is formed.

Accordingly, we propose
\textbf{Adaptive Identity Feature Recovery (AIFR)}, a framework that inserts a lightweight recovery module at a single intermediate layer of a frozen FR backbone.
The module predicts adaptive residual corrections, with sample-level strength controlling its overall
magnitude and spatially adaptive feature correction modulating its update distribution for each token. The corrected features pass through the remaining frozen layers to produce recovered embeddings. During training, fused identity supervision aligns these
embeddings with clean identity targets in the pretrained FR space, while identity-shift score guides the predicted correction strength and identity preservation. At inference, AIFR requires only the input portrait and one forward pass, without reference images. The recovered embeddings support direct matching against the unchanged gallery and compatibility with downstream embedding-conditioned
models (Fig.~\ref{application}(c)).




We evaluate AIFR on diffusion-based makeup and anime portraits
across makeup types, stylization strengths, and generators.
AIFR consistently improves identity matching, increasing TAR at FAR=$10^{-4}$
from $56.75\%$ to $88.07\%$ in
the makeup setting.
We further input the recovered embeddings into a pretrained
Arc2Face model,
improving identity preservation in the reconstructed faces.

Our contributions are summarized as follows:
\begin{itemize}
    \item We introduce AIFR, an identity feature recovery framework for recovering FR embeddings from diffusion-based makeup and anime portraits. It retains the pretrained FR backbone and stored gallery templates, without requiring reference images at inference time.

    \item We develop a lightweight plug-in feature correction module that supports adaptive feature correction at the intermediate layer of a frozen FR backbone, which combines sample-level strength prediction with spatially adaptive feature correction to recover identity features.

    \item We evaluate the compatibility of AIFR across makeup and anime settings, including unseen makeup types, stylization strengths, and generators. Conditioned on recovered FR embeddings, we perform downstream face reconstruction, showing compatibility for this embedding-conditioned downstream model.
\end{itemize}

\begin{figure*}
    \centering
    \includegraphics[width=1\linewidth]{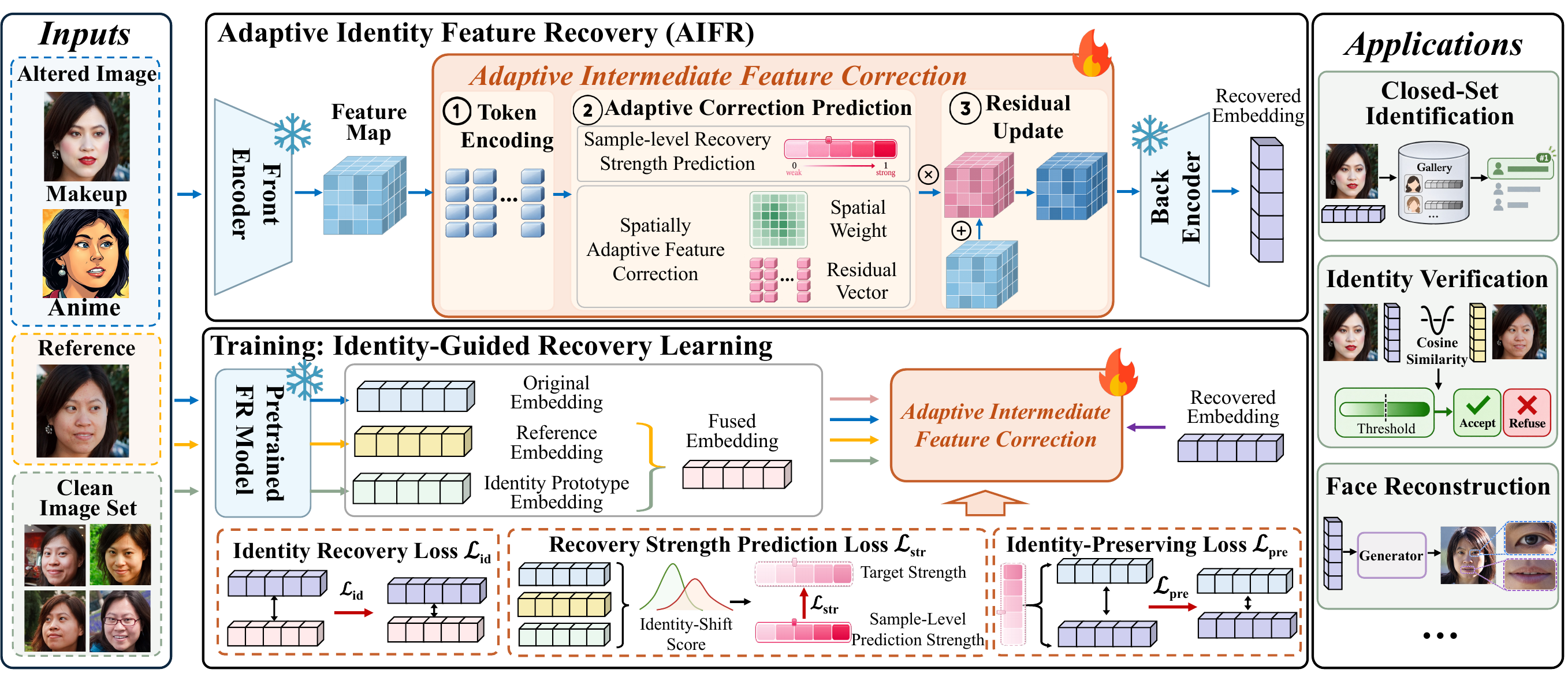}
    \caption{\textbf{Overview of AIFR Framework.} Given a makeup or anime portrait, the frozen FR front encoder first extracts an intermediate feature map. The trainable \textit{Adaptive Intermediate Feature Correction} module then predicts a residual correction in three steps: \ding{172} token encoding, \ding{173} adaptive correction prediction at the sample and spatial levels, and \ding{174} residual update. The corrected features are passed through the frozen FR back encoder to produce the recovered identity embedding. During training, identity-guided recovery learning optimizes only the recovery module. At inference, only the altered portrait is required, enabling direct closed-set identification and identity verification, as well as downstream face reconstruction. 
 }
    \label{pipeline}
\end{figure*}

\section{Related Work}
\label{sec:related_work}
\subsection{Identity-Preserving Face Generation/Editing}
Recent face generation and editing methods enable appearance changes while preserving the source identity. Makeup transfer has evolved from GAN-based and spatial correspondence methods~\cite{li2018beautygan,chen2019beautyglow,gu2019ladn,jiang2020psgan,nguyen2021lipstick,sun2022ssat,yang2022elegant,yan2023beautyrec} to diffusion-based approaches~\cite{zhang2025stablemakeup,zhu2025fluxmakeup}. In particular, Stable-Makeup~\cite{zhang2025stablemakeup} combines makeup-detail encoding with content and structure control for real-world makeup transfer, while FLUX-Makeup~\cite{zhu2025fluxmakeup} performs identity-consistent reference makeup transfer within a diffusion-transformer framework. For personalized face generation, recent methods inject identity information from one or more reference images into pretrained generative models to synthesize the same subject across diverse appearances and styles~\cite{infiniteyou2025iccv,wang2024instantid,guo2024pulid,li2024photomaker,xiao2025fastcomposer}. For example, InstantID~\cite{wang2024instantid} enables zero-shot personalization from a single facial image, and PhotoMaker~\cite{li2024photomaker} aggregates identity information from multiple reference images through stacked ID embeddings. 

These generation methods aim
to preserve identity, but heavy makeup and stylization still hinder identity matching to clean faces~\cite{wang2020fm2unet,2026styleid}. Our work focuses on recovering identity features from such diffusion-based
portraits. 



\subsection{Face Restoration}
Face restoration provides an image-level approach to reducing appearance differences before recognition. General blind face restoration methods, such as GFP-GAN~\cite{wang2021gfpgan}, RestoreFormer~\cite{wang2022restoreformer}, VQFR~\cite{gu2022vqfr}, and CodeFormer~\cite{zhou2022codeformer}, reconstruct high-quality facial images from inputs affected by low resolution, blur, noise, or other unknown degradations using learned facial priors. For makeup and beautification changes, restoration is more directly formulated as makeup or beautification removal. SAMC~\cite{samc2020} introduces semantic-aware makeup cleanser to remove facial makeup. DeBeauty~\cite{wang2025debeauty} targets facial beautification removal by restoring both appearance and structural changes. MAD~\cite{ruan2025mad} conducts makeup transfer, beauty filtering, and makeup removal within a unified cross-domain diffusion framework, while DiffClean~\cite{gavas2026diffclean} uses text-guided diffusion to remove makeup and evaluates the restored images for age estimation and face verification. 
Restored images should still be encoded into identity embeddings for recognition, and residual makeup or artifacts can distort the identity cues. In contrast, we correct intermediate features within a frozen FR model to recover the identity embedding used for recognition, avoiding this image restoration step. 




\subsection{Face Recognition}
Modern face recognition models learn discriminative face representations through large-scale training and margin-based objectives, such as ArcFace~\cite{deng2019arcface}, MagFace~\cite{meng2021magface}, AdaFace~\cite{2022adaface}, and ElasticFace~\cite{boutros2022elasticface}. Beyond general face recognition, several methods improve robustness under challenging imaging or appearance conditions. For example, MagFace~\cite{meng2021magface} and AdaFace~\cite{2022adaface} incorporate sample quality into representation learning to improve recognition of low-quality faces. 

For makeup variation, FM$^2$u-Net~\cite{wang2020fm2unet} develops a multi-branch architecture for makeup-invariant face verification, but we exclude it from our baselines as no official implementation is publicly available. For stylized faces, Stylized-Face~\cite{peng2025stylizedface} provides a large-scale dataset to study and improve recognition of stylized faces. StyleID~\cite{2026styleid} introduces perception-aware datasets and an identity metric for evaluating facial identity across stylization.
These approaches improve robustness by training or adapting new recognition models for particular challenging conditions, requiring gallery re-encoding when their output embeddings are incompatible with existing gallery templates~\cite{shen2020bct}.
More closely related to feature correction, Ozturk et al.~\cite{ozturk2025filters} study the effect of social-media filters on face recognition and 
use a separate embedding correction model for each known filter to improve recognition.
We instead consider diverse appearance changes in diffusion-based portraits. By conducting adaptive feature correction across facial regions, we recover identity embeddings in the original space, enabling direct matching against unchanged clean gallery templates.




\section{Method}

\subsection{Framework Overview}
Let $F(\cdot)$ denote a pretrained face recognition (FR) model that maps an aligned face image to an $\ell_2$-normalized identity embedding.
Given a diffusion-based makeup or anime portrait $x_i$, our goal is to recover identity features for reliable recognition while allowing the resulting embedding to be compared with existing gallery templates. Full-backbone fine-tuning without explicit compatibility constraints may change the FR embedding space, potentially requiring gallery re-encoding~\cite{shen2020bct}.
To adapt to appearance changes while retaining the pretrained
recognizer, we keep its layers fixed and learn corrections
through an inserted recovery module.
We train the module to align the final FR embeddings with
clean identity targets extracted by the original model, guiding recovery in the same space as the stored gallery templates.

Appearance changes can affect facial regions unevenly,
motivating corrections at individual spatial locations.
The intermediate feature map retains spatially organized facial
cues, allowing corrections at individual locations before the remaining frozen layers
produce the final identity embedding.
Motivated by this, we propose \textbf{Adaptive Identity Feature Recovery (AIFR)}, a framework that learns to correct identity features within a frozen FR model by inserting a lightweight \emph{Adaptive Intermediate Feature Correction}
module at a single intermediate layer of the frozen recognizer. The overview of AIFR is shown in Fig. \ref{pipeline}.

We split the pretrained FR model $F$ at the insertion layer into a front encoder $F_{\mathrm{f}}(\cdot)$ and a back encoder $F_{\mathrm{b}}(\cdot)$, i.e., $F(x)=F_{\mathrm b}(F_{\mathrm f}(x))$. The front encoder $F_{\mathrm{f}}$ extracts a spatial feature map, and the
back encoder $F_{\mathrm b}$ maps it to an $\ell_2$-normalized FR
embedding. For an appearance-altered portrait $x_i$, the recovery process is
\begin{equation}
    H_i=F_{\mathrm f}(x_i),
    \qquad
    \widehat z_i=F_{\mathrm b}(H_i+\Delta H_i),
    \label{eq:recovered_embedding}
\end{equation}
where $H_i\in\mathbb R^{C\times h\times w}$ denotes the intermediate feature map and $\Delta H_i$
is an input-dependent residual predicted by the recovery module. Both parts of $F$ remain frozen, and only the recovery module is
optimized in this way. 

As illustrated in Fig.~\ref{pipeline}, the recovery module
follows three steps: \emph{\ding{172} Token encoding} encodes the spatial features into
tokens that share information across facial regions,
\emph{\ding{173} Adaptive correction prediction} uses two parallel
branches to adapt corrections at the sample and spatial levels:
sample-level strength prediction controls the overall update
magnitude for each input, while spatially adaptive feature
correction predicts a spatial weight and a residual vector for each token, and \emph{\ding{174} Residual update} combines these predictions into
$\Delta H_i$, which is added to $H_i$ before the frozen back
encoder produces the recovered embedding.

\subsection{Adaptive Intermediate Feature Correction}
\label{sec:feature_correction}


\noindent\textbf{Step 1. Token encoding.}
To predict each local correction using context from other
facial regions, we encode the spatial features jointly. We reshape $H_i$ into $N=hw$ spatial tokens
$X_i\in\mathbb R^{N\times C}$ with one token per spatial
location, and add learnable positional embeddings
$P\in\mathbb R^{N\times C}$ to encode their locations.
A lightweight Transformer encoder $\Phi_\theta$ produces
\begin{equation}
    T_i = \Phi_{\theta} (X_i+P)\in\mathbb{R}^{N\times C}.
\end{equation}
The shared tokens $T_i$ combine information from multiple facial regions.

\noindent\textbf{Step 2. Adaptive correction prediction.}
To accommodate differences across inputs and facial regions,
we adapt corrections at the sample and spatial levels
in parallel.
Sample-level strength prediction controls the overall
correction magnitude, while spatially adaptive feature
correction specifies the correction at each location.


\noindent\emph{\textbf{Sample-level recovery strength prediction.}}
Appearance changes affect identity features to different degrees,  motivating an input-dependent correction magnitude.
We pool the tokens
and predict a sample-level
correction strength:
\begin{equation}
    \bar t_i=\frac{1}{N}\sum_{j=1}^{N}T_{i,j},
    \qquad
s_i=\sigma\!\left(\psi_s(\bar t_i)\right),
\end{equation}
where $\psi_s:\mathbb R^C\rightarrow\mathbb R$ is a learnable strength estimator and $\sigma$ is the sigmoid function.
The strength $s_i\in[0,1]$ controls the overall update of feature correction for each input image $x_i$, allowing stronger correction when needed and limiting unnecessary changes. We supervise this prediction using the identity-shift score defined
in Sec.~\ref{sec:recovery_learning}.


\noindent\emph{\textbf{Spatially adaptive feature correction.}}
Appearance changes can affect facial regions to different extents.
To control the correction at each location, we predict
a spatial weight and a residual vector for each token:
\begin{equation}
    M_i=\sigma\!\left(\psi_m(T_i)\right)\in[0,1]^{N\times1},
    \quad
    R_i=\psi_r(T_i)\in\mathbb R^{N\times C}.
\end{equation}
Here, $\psi_m:\mathbb R^C\rightarrow\mathbb R$ is a spatial
modulator that controls the correction applied to the $j$-th spatial token, and $\psi_r:\mathbb R^C\rightarrow\mathbb R^C$ is a
residual predictor. Both mappings are applied to individual tokens with parameters shared
across spatial locations.
For token $j$, $R_{i,j}$ specifies its feature correction vector, while
$M_{i,j}$ controls how strongly that vector contributes to the update.

\noindent\textbf{Step 3. Residual update.}
With the sample strength, spatial weight, and residual vectors for each token, the final correction is formulated as
\begin{equation}
    \Delta X_i= s_i
\left(
M_i\odot\widetilde R_i
\right), 
\end{equation}
where $\widetilde R_i=
\frac{\beta R_i}
{
\max\left(
\sqrt{\frac{1}{NC}\lVert R_i\rVert_F^2},
\epsilon
\right)
}$ is normalized using a sample-wise root-mean-square statistic~\cite{RMS2019} so that the strength predictor
controls updates on a consistent scale,  and $\odot$ denotes element-wise multiplication,
with $M_i$ broadcast across channels. Finally, $\Delta X_i$ is reshaped into $\Delta H_i\in\mathbb R^{C\times h\times w}$ and is applied as in Eq.~\ref{eq:recovered_embedding}.

\subsection{Identity-Guided Recovery Learning}
\label{sec:recovery_learning}
We train the recovery module to correct the intermediate feature map $H_i$, guided by identity-related losses on the final FR embeddings.

\noindent \textbf{Fused identity supervision.} 
A paired clean reference provides a target corresponding to each altered sample, while an identity prototype aggregates clean
images of the same person, reducing dependence on a single
reference. We combine these cues to guide identity recovery.


For an appearance-altered image $x_i$ of identity $y$, we use a paired clean reference $x_i^{r}$ and a clean image set $\mathcal C_y$ from the same identity during training. Their reference FR embedding and the identity prototype can be computed as
\begin{equation}
    z_i^{r}= F(x_i^{r}), 
    \qquad
c_y=\frac{\sum_{x\in\mathcal C_y}F(x)}{\left\|\sum_{x\in\mathcal C_y}F(x)\right\|_2}.
\end{equation}
By combining these two identity cues into a fused target, we define the identity recovery loss that maximizes its cosine similarity with the recovered FR embedding: 
\begin{equation}
\mathcal{L}_{\mathrm{id}}=w_i\left[
1-\cos\left(
\widehat z_i,
\alpha z_i^{r}+(1-\alpha)c_y
\right)
\right],
\end{equation}
where $w_i>0$ is a sample weight based on
the normalized identity-shift score and sample difficulty, and $\alpha\in[0,1]$ controls their relative contribution, encouraging the recovered FR embedding $\widehat z_i$ to match the clean identity target.
The normalization rules and sample weighting are
detailed in Appendix~A.1.
Note that the recovery process is supervised in the original FR embedding space, as both identity cues are obtained by the original FR model. 

\noindent \textbf{Identity-shift supervision.}
The required correction depends on how appearance changes
affect the FR identity features.
We therefore derive an explicit target strength from the deviation
of the original embedding from clean identity cues. Let $z_i=F_{\mathrm b}(H_i)$ denote the original FR embedding for input image $x_i$ before recovery.
We define its identity-shift score as
\begin{equation}
d_i=
\frac{1}{2}
\max\left(
1-\cos\left(z_i,z_i^{r}\right)
+\cos\left(z_i^{r},c_y\right)
-\cos\left(z_i,c_y\right),0
\right).
\end{equation}
The first term (i.e.,$1-\cos\left(z_i,z_i^{r}\right)$) measures the deviation from the paired clean reference, and the second term (i.e., $\cos\left(z_i^{r},c_y\right)
-\cos\left(z_i,c_y\right)$) measures the decrease in
prototype similarity relative to that reference.

We normalize $d_i$ to get the target recovery strength $s_i^{*}\in[0,1]$, and constrain the predicted recovery strength $s_i$ toward this target. The recovery strength prediction loss is
\begin{equation}
    \mathcal L_{\mathrm{str}}
    =
    \left\|s_i-s_i^{*}\right\|_2^2.
\end{equation}

\noindent\textbf{Identity preservation.}
To limit unnecessary changes for inputs with little identity shift, we use an identity-preserving loss:
\begin{equation}
   \mathcal L_{\mathrm{pre}}= (1-s_i^{*})
\left[
1-\cos\left(\widehat z_i,z_i\right)
\right].
\end{equation}
For inputs with small identity shifts (i.e., small $s_i^*$), this loss keeps the
recovered embedding close to the original embedding.
Its weight decreases as the target strength increases, allowing greater correction toward the clean identity.

The complete training objective is
\begin{equation}
\mathcal L= \mathcal L_{\mathrm{id}}
+
\lambda_{\mathrm{str}}\mathcal L_{\mathrm{str}}
+
\lambda_{\mathrm{pre}}\mathcal L_{\mathrm{pre}},
\end{equation}
where $\lambda_{\mathrm{str}}\geq0$ and $\lambda_{\mathrm{pre}}\geq0$
are weights for the corresponding loss terms. Gradients pass through the frozen back encoder to update only
the recovery module.
The paired reference and clean images are used only for training supervision. During inference, the recovery module requires only the appearance-altered image as an input to obtain the recovered FR embeddings for identity matching.


\subsection{Applications of Recovered Embeddings}
\label{sec:embedding_applications}


\noindent\textbf{Closed-set identification.}
The recovered embedding $\widehat z_i$ is compared with existing gallery embeddings to retrieve the identity that best matches the input portrait. This comparison uses the original gallery templates without re-encoding.

\noindent\textbf{Identity verification.}
Given a claimed identity, $\widehat z_i$ is compared with its stored reference embedding to determine whether the input portrait depicts the same identity.

\noindent\textbf{Face reconstruction.}
Beyond recognition, we evaluate whether the recovered embedding remains compatible with embedding-conditioned downstream models. Specifically, $\widehat z_i$ is directly fed into the pretrained Arc2Face model~\cite{papantoniou2024arc2face} for face reconstruction, and the generated faces are re-encoded only to quantify identity preservation.

\section{Experiments}

\begin{figure}
    \centering
\includegraphics[width=1\linewidth]{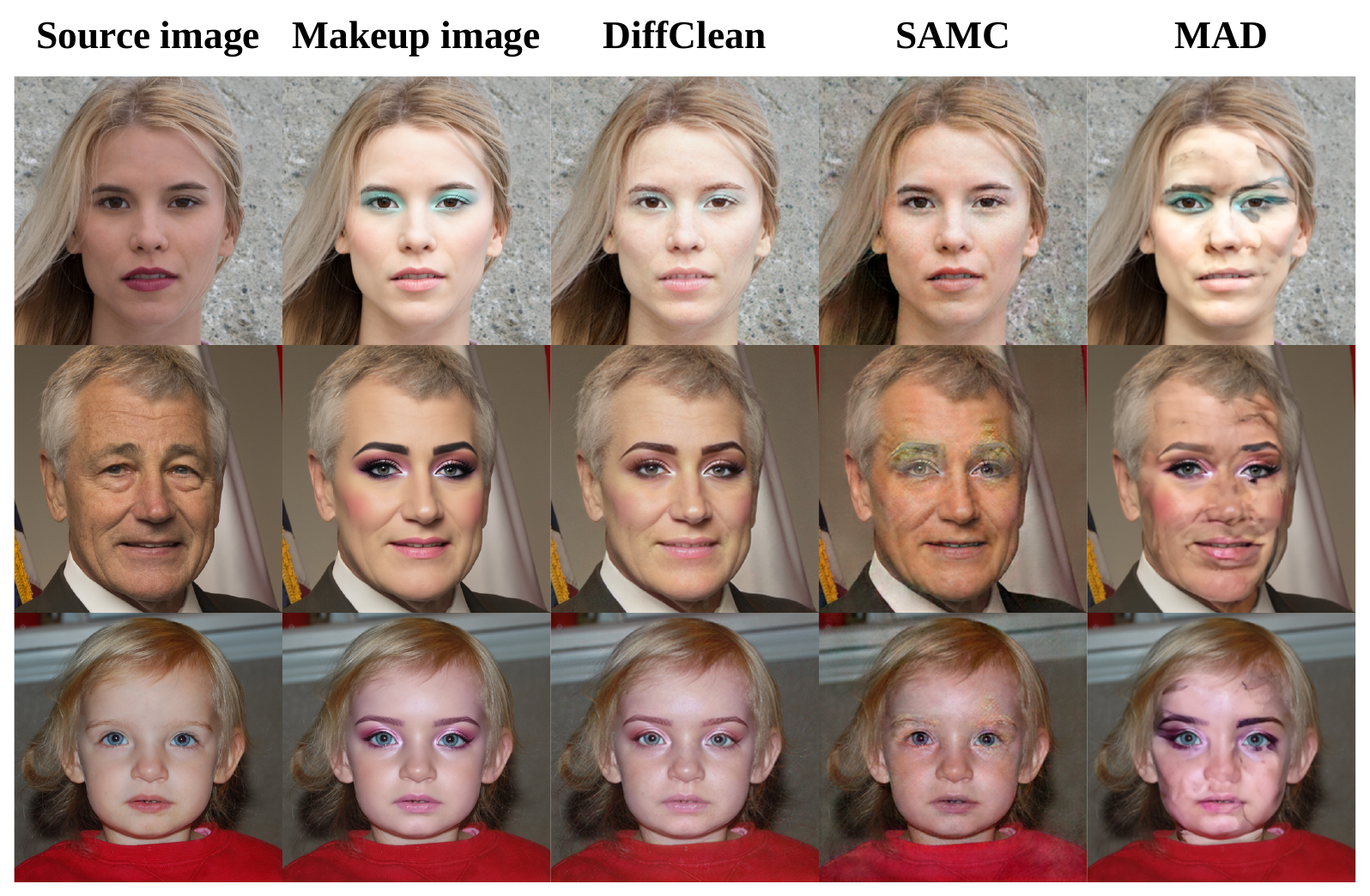}
    \caption{Makeup removal examples on FFHQ.}
   \label{makeup_removal}
\end{figure}

\subsection{Experimental Setup}

\subsubsection{Datasets}
We construct diffusion-based makeup portraits from FFHQ~\cite{FFHQ2019} and anime portraits from VGGFace2~\cite{cao2018vggface2}.


\noindent\textbf{Training.}
We use FLUX-Makeup~\cite{zhu2025fluxmakeup} to synthesize makeup training images from FFHQ, including 47002 makeup images  from 5734 identities across three makeup types. During model selection, 5161 identities are used for inner training, and 573 identities are used for validation. For anime, we use InstantID~\cite{wang2024instantid} to generate 7200 images from 400 VGGFace2 identities across three stylization strengths, with a $9{:}1$ identity split for inner training and validation.



\noindent\textbf{Evaluation.}
We test held-out 100 FFHQ and 50 VGGFace2 identities for makeup and anime, respectively. The main test sets contain 1800 makeup and 900 anime images, with six images per identity for each of the three makeup types or stylization strengths. Generalization is evaluated on 1200 makeup images across two unseen types and 1200 anime images across four unseen strengths. Cross-method evaluation uses Stable-Makeup~\cite{zhang2025stablemakeup} for makeup and 5075 anime images from PhotoMaker~\cite{li2024photomaker}, PuLID~\cite{guo2024pulid}, and InfiniteYou~\cite{infiniteyou2025iccv}. Note that these generators are not used for training. The identification galleries contain 152 identities for makeup (including all 100 test identities) and all 50 test identities for anime.

\subsubsection{Metrics}
We evaluate identity recovery using closed-set identification and identity verification metrics. For identification, we report \textbf{Rank-1}  accuracy by comparing each test image with clean identity centers using cosine similarity.
For verification, we report \textbf{True Accept Rate (TAR)} at a fixed false accept rate (FAR), i.e., FAR=$10^{-3}$ and $10^{-4}$. Following the standard face verification protocol~\cite{2022adaface}, each recognizer has its own thresholds calibrated at FAR=$10^{-3}$ and $10^{-4}$ using clean impostor comparisons. The calibration sets contain 1283 images from 445 FFHQ identities and 3300 images from 550 VGGFace2 identities, respectively, and are identity-disjoint from all evaluations. Thresholds remain fixed when testing altered faces. TAR is the fraction of genuine pairs exceeding each threshold. The detailed evaluation protocols are provided in Appendix A.3.

For the downstream face reconstruction task, we further report
\textbf{DINOv2 feature similarity}~\cite{oquab2023dinov2} to evaluate the visual consistency between each reconstructed face and its clean source. We extract global CLS features using DINOv2 ViT-B/14 and compute cosine similarity. 

\subsubsection{Implementation Details}
We use the InsightFace AntelopeV2 ArcFace recognizer employed by Arc2Face~\cite{papantoniou2024arc2face}. Faces are aligned using five facial landmarks and resized to $112\times112$. AIFR applies the intermediate feature map of size $256\times14\times14$ and contains approximately 1.22M trainable parameters. The remaining ArcFace layers remain frozen and produce a 512-dimensional $\ell_2$-normalized FR embedding.

The makeup and anime recovery modules are trained independently from random initialization. We use identity-disjoint inner training and validation splits for model selection, with at most 50 epochs, a patience of five epochs, and a minimum improvement of $10^{-4}$. After the best epoch is determined, the model is reset to its initial state and retrained on the complete training pool for the selected number of epochs. The selected training lengths are six epochs for makeup recovery and three epochs for anime recovery. All recovery models are optimized with AdamW using a learning rate of $10^{-4}$, weight decay of $10^{-4}$, batch size $64$, and gradient clipping at 1.0. We set $\lambda_{\mathrm{str}}=0.5$, $\lambda_{\mathrm{pre}}=0.25$, and $\alpha=0.3$ for loss functions. Training is performed on a single NVIDIA H100 NVL GPU. More experimental details are provided in Appendix A.1.

\subsubsection{Baselines} 
We compare AIFR with different FR models and face restoration baselines. ArcFace~\cite{deng2019arcface} directly recognizes the altered images using the same pretrained FR model employed by AIFR and serves as the primary reference. AdaFace~\cite{2022adaface} provides an additional quality-aware FR baseline. We also fine-tune the complete pretrained ArcFace backbone on the same training data as AIFR using the ArcFace identity-classification loss, denoted as ArcFace-FT, to compare feature recovery with full-model adaptation.

For makeup removal, we evaluate SAMC~\cite{samc2020}, MAD~\cite{ruan2025mad}, and DiffClean~\cite{gavas2026diffclean}. These methods first remove the makeup appearance, after which the restored image is processed by the original ArcFace recognizer. For anime faces, we additionally compare with StyleID~\cite{2026styleid},  an identity encoder designed for recognition under stylization.

\begin{table}[t]
\centering
\caption{
Comparison on diffusion-based makeup and anime faces generated by FLUX-Makeup~\cite{zhu2025fluxmakeup} and Instantid~\cite{wang2024instantid}, respectively. Best and second-best results are shown in \textbf{bold} and \underline{underline}.
}
\label{main_results}
\footnotesize
\setlength{\tabcolsep}{5.5pt}
\renewcommand{\arraystretch}{1.2}
\begin{tabular}{@{}llccc@{}}
\toprule
Setting & Method &
Rank-1 $\uparrow$ &
TAR@$10^{-3}$ $\uparrow$ &
TAR@$10^{-4}$ $\uparrow$ \\
\midrule

\multirow{7}{*}{Makeup}
& ArcFace~\cite{deng2019arcface}
& 87.00 & 75.47 & 56.75 \\
& AdaFace~\cite{2022adaface}
& 79.67 & 63.47 & 36.88 \\
& ArcFace-FT
& 88.83 & 64.72 & 47.33 \\
\addlinespace[1.5pt]
& DiffClean~\cite{gavas2026diffclean}
& 85.72 & 73.28 & 53.94 \\
& SAMC~\cite{samc2020}
& \underline{93.28} & \underline{87.53} & \underline{74.78} \\
& MAD~\cite{ruan2025mad}
& 73.94 & 51.25 & 24.47 \\
\addlinespace[1.5pt]
& \textbf{AIFR}
& \textbf{96.11} & \textbf{93.72} & \textbf{88.07} \\

\midrule

\multirow{5}{*}{Anime}
& ArcFace~\cite{deng2019arcface}
& 94.89 & 85.15 & 75.46 \\
& AdaFace~\cite{2022adaface}
& 93.67 & 83.93 & 73.26 \\
& ArcFace-FT
& \textbf{97.22} & \underline{91.69} & \underline{77.87} \\
\addlinespace[1.5pt]
& StyleID~\cite{2026styleid}
& 96.33 & 65.39 & 37.54 \\
\addlinespace[1.5pt]
& \textbf{AIFR}
& \underline{96.78} & \textbf{95.33} & \textbf{92.52}\\

\bottomrule
\end{tabular}
\end{table}

\begin{table}[t]
\centering
\caption{
Generalization on makeup faces.
Cross-Makeup evaluates two unseen makeup types generated by FLUX-Makeup~\cite{zhu2025fluxmakeup},
while Cross-Method evaluates makeup faces generated
by the unseen Stable-Makeup model~\cite{zhang2025stablemakeup}.
}
\label{tab:makeup_generalization}
\footnotesize
\setlength{\tabcolsep}{5.5pt}
\renewcommand{\arraystretch}{1.2}
\begin{tabular}{@{}llccc@{}}
\toprule
Setting & Method &
Rank-1 $\uparrow$ &
TAR@$10^{-3}$ $\uparrow$ &
TAR@$10^{-4}$ $\uparrow$ \\
\midrule

\multirow{7}{*}{\shortstack{Cross-\\Makeup}}
& ArcFace~\cite{deng2019arcface}
& 89.50 & 79.50 & 60.58 \\
& AdaFace~\cite{2022adaface}
& 83.17 & 66.73 & 37.81 \\
& ArcFace-FT
& 88.92 & 64.69 & 46.73 \\
\addlinespace[1.5pt]
& DiffClean~\cite{gavas2026diffclean}
& 88.58 & 77.92 & 57.75 \\
& SAMC~\cite{samc2020}
& \underline{94.33} & \underline{89.38} & \underline{79.00} \\
& MAD~\cite{ruan2025mad}
& 76.33 & 53.50 & 25.79 \\
\addlinespace[1.5pt]
& \textbf{AIFR}
& \textbf{95.83} & \textbf{92.92} & \textbf{85.69} \\

\midrule

\multirow{7}{*}{\shortstack{Cross-\\Method}}
& ArcFace~\cite{deng2019arcface}
& 90.67 & 69.71 & 40.88 \\
& AdaFace~\cite{2022adaface}
& 76.00 & 47.46 & 20.04 \\
& ArcFace-FT
& 90.00 & 50.83 & 35.79 \\
\addlinespace[1.5pt]
& DiffClean~\cite{gavas2026diffclean}
& 89.33 & 65.96 & 36.67 \\
& SAMC~\cite{samc2020}
& \underline{91.83} & \underline{72.17} & \underline{43.38} \\
& MAD~\cite{ruan2025mad}
& 73.83 & 39.29 & 12.63 \\
\addlinespace[1.5pt]
& \textbf{AIFR}
& \textbf{93.00} & \textbf{77.29} & \textbf{51.54} \\

\bottomrule
\end{tabular}
\end{table}

\begin{table}[t]
\centering
\caption{
Generalization on anime-stylized faces.
Cross-Strength evaluates unseen stylization strengths.
Cross-Method evaluates the unseen personalization
methods, including PhotoMaker~
\cite{li2024photomaker}, PuLID~\cite{guo2024pulid}, and InfiniteYou~\cite{infiniteyou2025iccv}.
}
\label{tab:anime_generalization}
\footnotesize
\setlength{\tabcolsep}{5.5pt}
\renewcommand{\arraystretch}{1.2}
\begin{tabular}{@{}llccc@{}}
\toprule
Setting & Method &
Rank-1 $\uparrow$ &
TAR@$10^{-3}$ $\uparrow$ &
TAR@$10^{-4}$ $\uparrow$ \\
\midrule

\multirow{5}{*}{\shortstack{Cross-\\Strength}}
& ArcFace~\cite{deng2019arcface}
& 98.67 & 95.26 & \underline{90.65} \\
& AdaFace~\cite{2022adaface}
& 97.50 & 94.62 & 90.15 \\
& ArcFace-FT
& \underline{98.75} & \underline{96.65} & 89.19 \\
\addlinespace[1.5pt]
& StyleID~\cite{2026styleid}
& \textbf{98.92} & 79.01 & 51.78 \\
\addlinespace[1.5pt]
& \textbf{AIFR}
& \underline{98.75} & \textbf{98.14} & \textbf{97.15} \\

\midrule

\multirow{5}{*}{\shortstack{Cross-\\Method}}
& ArcFace~\cite{deng2019arcface}
& 78.74 & \underline{64.17} & \underline{52.25} \\
& AdaFace~\cite{2022adaface}
& 76.82 & 60.21 & 47.68 \\
& ArcFace-FT
& 74.71 & 52.13 & 32.25 \\
\addlinespace[1.5pt]
& StyleID~\cite{2026styleid}
& \textbf{81.27} & 46.71 & 25.10 \\
\addlinespace[1.5pt]
& \textbf{AIFR}
& \underline{80.51} & \textbf{70.86} & \textbf{60.20}\\

\bottomrule
\end{tabular}
\end{table}

\begin{table*}[t]
\centering
\caption{
Identity preservation and visual consistency of Arc2Face reconstructions
conditioned on original, ArcFace-FT, and AIFR-recovered embeddings.
All methods are evaluated using ArcFace~\cite{deng2019arcface},
AdaFace~\cite{2022adaface}, and MagFace~\cite{meng2021magface}. DINOv2 sim. denotes visual consistency with the clean source. R1 denotes Rank-1 accuracy. TAR$_3$ and TAR$_4$ denote TAR at thresholds at FAR=$10^{-3}$ and FAR=$10^{-4}$.
}
\label{tab:arc2face_reconstruction}

\small
\setlength{\tabcolsep}{6pt}
\renewcommand{\arraystretch}{1.0}

{
\begin{tabular}{ll ccc ccc ccc c}
\toprule

\multirow{2}{*}{Setting} &
\multirow{2}{*}{Method} &
\multicolumn{3}{c}{ArcFace} &
\multicolumn{3}{c}{AdaFace} &
\multicolumn{3}{c}{MagFace} &
\multirow{2}{*}{DINOv2 Sim. $\uparrow$}
\\

\cmidrule(lr){3-5}
\cmidrule(lr){6-8}
\cmidrule(lr){9-11}

& &
R1 $\uparrow$ &
TAR$_{3}$ $\uparrow$&
TAR$_{4}$ $\uparrow$&
R1 $\uparrow$ &
TAR$_{3}$ $\uparrow$ &
TAR$_{4}$ $\uparrow$ &
R1 $\uparrow$ &
TAR$_{3}$ $\uparrow$ &
TAR$_{4}$ $\uparrow$ &
\\

\midrule

\multirow{3}{*}{\shortstack{Makeup}}
& w/o AIFR
& 80.59 & 66.97 & 48.25
& 74.78 & 58.31 & 34.95
& 72.92 & 57.54 & 35.39
& 0.3840
\\

& ArcFace-FT
& 8.77 & 1.07 & 0.03
& 4.93 & 0.33 & 0.00
& 5.70 & 0.52 & 0.00
& 0.2253
\\

& \textbf{AIFR}
& \textbf{93.31} & \textbf{88.73} & \textbf{74.56}
& \textbf{89.47} & \textbf{81.58} & \textbf{56.96}
& \textbf{89.58} & \textbf{81.20} & \textbf{56.11}
& \textbf{0.4899}
\\

\midrule

\multirow{3}{*}{\shortstack{Anime}}
& w/o AIFR
& 64.67 & 27.93 & 9.87
& 60.33 & 24.73 & 10.87
& 57.67 & 26.40 & 10.07
& 0.3081
\\

& ArcFace-FT
& 19.33 & 2.93 & 0.20
& 15.67 & 2.07 & 0.13
& 15.00 & 1.47 & 0.20
& 0.2062
\\

& \textbf{AIFR}
& \textbf{87.67} & \textbf{74.40} & \textbf{56.80}
& \textbf{86.33} & \textbf{72.87} & \textbf{52.87}
& \textbf{86.00} & \textbf{72.47} & \textbf{50.87}
& \textbf{0.4030}
\\

\bottomrule
\end{tabular}
}

\end{table*}

\begin{figure}
    \centering
\includegraphics[width=1\linewidth]{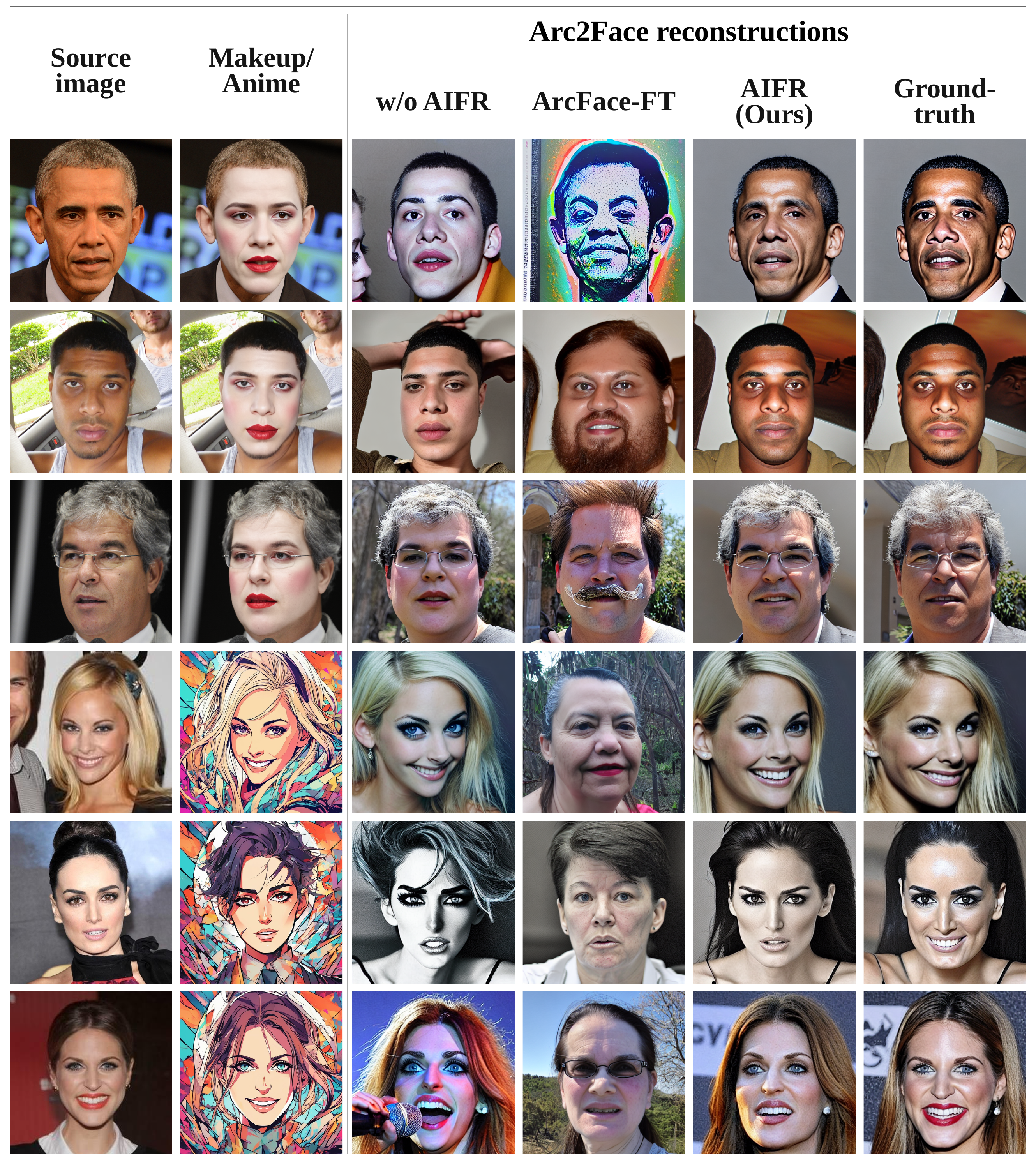}
    \caption{Qualitative comparison of Arc2Face reconstructions from the evaluated FR embeddings.}
    \label{reconstruction}
\end{figure}
\subsection{Main Comparison}

Table~\ref{main_results} reports the main results on diffusion-based makeup and anime faces. For Makeup, AIFR achieves the highest scores across all three metrics, with 96.11\% Rank-1 accuracy. At FAR=$10^{-4}$, it improves TAR from 56.75\% to 88.07\% over the ArcFace baseline and exceeds SAMC, the strongest competing method, by 13.29 percentage points.
For makeup removal baselines, SAMC improves recognition, whereas DiffClean and MAD score below the original ArcFace baseline. Fig.~\ref{makeup_removal} provides the qualitative comparisons, showing that these methods leave residual makeup and generate some artifacts.

For Anime, AIFR achieves the highest verification performance, reaching 95.33\% and 92.52\% TAR at FAR=$10^{-3}$ and $10^{-4}$, respectively. Compared with ArcFace-FT, these correspond to gains of 3.64 and 14.65 percentage points, although AIFR has 0.44 percentage points lower Rank-1 accuracy. Identification depends on the relative ranking of identity similarities, whereas verification measures genuine acceptance at a strict threshold. Therefore, the results highlight AIFR's advantage for verification at low FAR, alongside competitive identification accuracy. 
Further analysis of genuine and impostor similarity distributions under fixed verification thresholds is provided in Appendix~A.3.

\subsection{Generalization Analysis}
\noindent\textbf{Generalization to unseen makeup types and methods.} 
Table~\ref{tab:makeup_generalization} shows that AIFR achieves the best results across all three metrics in both settings. On Cross-Makeup, its TAR at FAR=$10^{-4}$ exceeds SAMC by 6.69 percentage points. On Cross-Method, AIFR reaches 93.00\% Rank-1 accuracy and 77.29\% TAR at FAR=$10^{-3}$, outperforming other baselines. These results indicate that AIFR transfers effectively across different makeup types and generation methods. 

\noindent\textbf{Generalization to unseen stylization strengths and methods.} Table~\ref{tab:anime_generalization} evaluates the anime faces under unseen stylization strengths and unseen personalization methods. AIFR obtains the highest verification scores in both settings. StyleID achieves slightly higher Rank-1 accuracy, but substantially lower verification scores. ArcFace-FT degrades significantly under unseen personalization methods, indicating weaker robustness to generator shifts. In contrast, AIFR preserves the frozen FR space and transfers more consistently across different generation methods.

\subsection{Downstream Compatibility}

To validate the utility and compatibility of the recovered FR embeddings for downstream tasks, we use the pretrained embedding-conditioned Arc2Face generator for face reconstruction~\cite{papantoniou2024arc2face}. We compare reconstructions conditioned on the original altered-face embeddings (\textit{w/o AIFR}), embeddings from ArcFace-FT, and AIFR-recovered embeddings. Identity preservation is evaluated with ArcFace, AdaFace, and MagFace, while DINOv2 similarity measures visual consistency with the corresponding clean source. 

In Table~\ref{tab:arc2face_reconstruction}, AIFR consistently improves the identity preservation of the reconstructed faces across both settings, indicating improved identity consistency across independent FR models. Fig.~\ref{reconstruction} provides qualitative comparisons. Reconstructions conditioned on the original altered embeddings (i.e., w/o AIFR) can exhibit noticeable identity drift, whereas conditioning on recovered embeddings (i.e., AIFR) produces faces with better preserved identity characteristics (e.g., Beard and skin tone). ArcFace-FT yields poor reconstructed images and scores despite its strong in-domain identification performance, which is consistent with a mismatch between its adapted embeddings and the conditioning space expected by the Arc2Face generator. Therefore, AIFR recovers identity information that remains usable by downstream embedding-conditioned applications without modifying the original recognition space, maintaining compatibility with an external generative model.





\begin{table}[t]
\centering
\caption{
Ablation of full fine-tuning and AIFR insertion position under
Makeup and Anime settings.
}
\label{tab:insertion_level}
\footnotesize
\setlength{\tabcolsep}{2.8pt}
\renewcommand{\arraystretch}{1.08}
\begin{tabular}{clccc}
\toprule
Setting
& Adaptation Strategy
& R1 $\uparrow$
& TAR@$10^{-3}$ $\uparrow$
& TAR@$10^{-4}$ $\uparrow$ \\
\midrule

\multirow{6}{*}{Makeup}
& Full fine-tuning
& 90.83 & 68.67 & 42.87 \\
\cmidrule(lr){2-5}
& $56{\times}56$
& 90.75 & 74.16 & 53.95 \\
& $28{\times}28$
& 90.75 & 74.96 & 54.93 \\
& \cellcolor{gray!10}$\mathbf{14{\times}14}$
& \cellcolor{gray!10}\textbf{91.41}
& \cellcolor{gray!10}\textbf{76.95}
& \cellcolor{gray!10}\textbf{57.16} \\
& $7{\times}7$
& 90.94 & 76.15 & 55.04 \\
& Pre-embedding
& 90.72 & 75.69 & 54.42 \\

\midrule

\multirow{6}{*}{Anime}
& Full fine-tuning
& 81.00 & 49.72 & 27.16 \\
\cmidrule(lr){2-5}
& $56{\times}56$
& 83.92 & 71.16 & 58.94 \\
& $28{\times}28$
& 85.29 & 74.68 & 63.88 \\
& \cellcolor{gray!10}$\mathbf{14{\times}14}$
& \cellcolor{gray!10}\textbf{85.75}
& \cellcolor{gray!10}\textbf{76.78}
& \cellcolor{gray!10}\textbf{65.65} \\
& $7{\times}7$
& 84.04 & 73.94 & 61.52 \\
& Pre-embedding
& 85.00 & 73.56 & 61.35 \\

\bottomrule
\end{tabular}
\end{table}

\subsection{Ablation Studies}
We conduct ablation studies on the same appearance-altered dataset covering two diffusion-based makeup generators and four anime stylization generators, with different makeup types and stylization strengths.

We ablate the recovery architecture by comparing full-backbone fine-tuning with feature recovery at different insertion positions under the same training objectives. As shown in Table~\ref{tab:insertion_level}, insertion at the $14{\times}14$ feature map performs best in both settings, outperforming pre-embedding correction by 2.74 and 4.30 percentage points in TAR at FAR=$10^{-4}$ for makeup and anime, respectively. This suggests that intermediate features provide more effective recovery cues than the final representation, while retaining spatial information for localized corrections. AIFR also outperforms full-backbone fine-tuning, highlighting the benefit of correcting features within a frozen FR backbone. Further ablations of the training objectives, recovery designs, and hyperparameters, and feature correction
visualizations and analysis, are provided in Appendix B.

\section{Conclusion}
We presented AIFR, an identity feature recovery framework for diffusion-based makeup and anime portraits. AIFR applies adaptive residual corrections at a single intermediate layer of a frozen FR backbone, controlling their strength and spatial distribution for each input. Fused identity supervision provides an alignment target, while identity-shift supervision regulates the recovery strength. Experiments across multiple makeup styles, stylization strengths, and generators demonstrate consistent improvements in recognition performance, with additional benefits for downstream face reconstruction. These findings support adaptive intermediate feature correction as an effective approach to identity recovery while allowing existing gallery templates to be reused. AIFR requires only a single forward pass and no reference images at inference, facilitating its integration into existing FR pipelines.

\section*{LLM Usage Statement}
OpenAI ChatGPT was used to assist with language refinement and figure presentation. All scientific content, experimental results, and conclusions were reviewed and verified by the authors.

{
    \small
    \bibliographystyle{ieeenat_fullname}
    \bibliography{main}
}

\appendix
\twocolumn[
\begin{center}
    {\LARGE\bfseries Appendix}
\end{center}
\vspace{6pt}
]

\section{Additional Experimental Details}
\label{app:experimental_details}

\subsection{Recovery Implementation}
\label{app:recovery_training}
The recovery module processes 196 spatial tokens with 256 channels using two Transformer blocks, eight attention heads, an MLP ratio of two, an adapter
bottleneck of 32, and dropout of 0.2.
The insertion interface is the output of \texttt{layer4.0.bn1} in the converted
AntelopeV2 backbone, corresponding to the $256\times14\times14$ feature map. The ArcFace backbone remains frozen, and the residual scale $\beta$ is fixed at 0.05. Only the recovery module is optimized, with 1,221,826 trainable parameters.

For strength supervision, the identity shift score is divided by its 90th percentile over the training pairs and clipped to $[0,1]$. The resulting target supervises the strength predictor and weights the preservation loss. The identity loss uses the sample weight
$w_i=(0.25+0.75s_i^*)\,\omega_i$,
where $\omega_i\in[1,1.5]$ is the difficulty factor. $\omega_i$ increases linearly as the cosine similarity between the appearance-altered input and clean reference decreases.
The main recognition and reconstruction checkpoints use $\alpha=0.3$, $\lambda_{\mathrm{str}}=0.5$, and $\lambda_{\mathrm{pre}}=0.25$, with six training epochs for Makeup and three for Anime. Clean references and identity prototypes are used for training supervision, and inference requires only the altered query.

\subsection{Baseline Implementation}
\label{app:baseline_details}
ArcFace-FT starts from the same pretrained ArcFace backbone and uses the same paired training data as AIFR.
For each pair, the altered and clean images contribute equally to an ArcFace classification objective with angular margin 0.5 and scale 64. The learning rates are $10^{-5}$ for Makeup and $10^{-4}$ for Anime. Both gallery embeddings and verification thresholds are recomputed for the fine-tuned embedding space.

SAMC is locally reproduced using a separate paired makeup dataset, and other baselines use pretrained checkpoints without additional training. We apply the respective preprocessing for each recognizer: ArcFace uses RGB inputs normalized to $[-1,1]$, AdaFace uses BGR inputs normalized to $[-1,1]$, MagFace uses BGR inputs in $[0,1]$, and StyleID uses a CLIP image processor at $224\times224$. Center cropping is used as a fallback when face detection fails.

\subsection{Evaluation Protocols}
\label{app:evaluation_details}
For each recognizer, verification thresholds are calibrated using clean impostor comparisons from 1,283 images of 445 FFHQ identities and 3,300 images of 550 VGGFace2 identities, excluding evaluation identities. At a target false acceptance rate (FAR) $r$, the threshold is the empirical $(1-r)$ quantile of impostor scores with the higher-quantile rule, and a match is accepted when its cosine similarity score meets or exceeds this threshold. Table~\ref{tab:verification_thresholds} lists the resulting thresholds for each recognizer. The Thresholds remain fixed for appearance-altered inputs. AIFR and the restoration methods followed by ArcFace reuse the original ArcFace thresholds, whereas ArcFace-FT and StyleID are calibrated in their respective embedding space.

\begin{figure*}[t]
    \centering
    \includegraphics[width=0.95\textwidth]{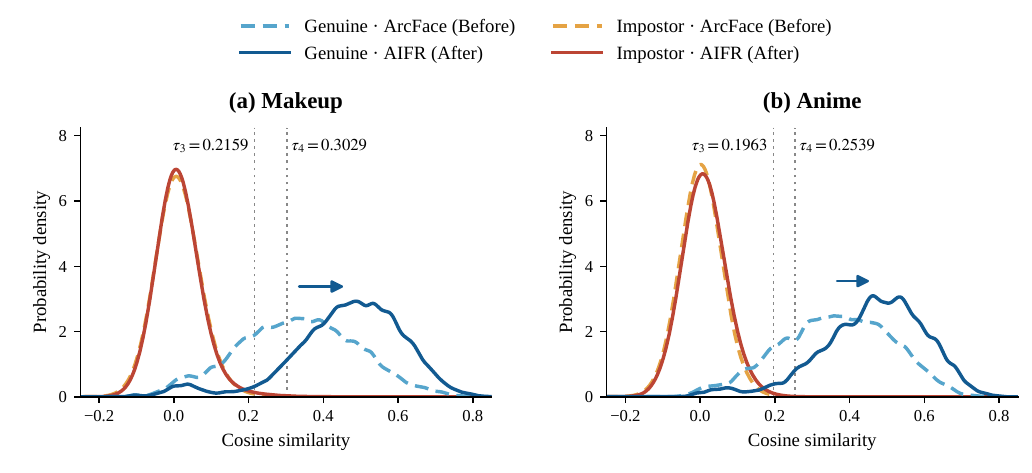}
    \caption{
    Genuine and impostor cosine similarity distributions before (ArcFace) and after AIFR on (a) Makeup and (b) Anime test sets. The vertical dashed lines indicate the fixed clean-calibrated thresholds at FARs of $10^{-3}$ ($\tau_3$) and $10^{-4}$ ($\tau_4$).
    }
    \label{fig:score_distributions}
\end{figure*}

Fig.~\ref{fig:score_distributions} further illustrates the verification results under these fixed thresholds. AIFR shifts the genuine similarity distributions toward higher scores, while the impostor distributions remain largely unchanged in both settings. This supports improved genuine matching without substantial changes to the overall impostor score distributions.

For downstream reconstruction, Arc2Face uses 25 denoising steps and a guidance scale of 3. Reconstructed images are evaluated with fixed ArcFace, AdaFace, and MagFace recognizers, and failed detections count as rejected matches. 
\begin{table}[b]
\centering
\caption{
Verification thresholds calibrated on clean
impostor pairs from FFHQ and VGGFace2
at target FARs of $10^{-3}$ and $10^{-4}$.
}
\label{tab:verification_thresholds}
\footnotesize
\setlength{\tabcolsep}{3.5pt}
\renewcommand{\arraystretch}{1.15}
\begin{tabular}{lcccc}
\toprule
& \multicolumn{2}{c}{FFHQ}
& \multicolumn{2}{c}{VGGFace2} \\
\cmidrule(lr){2-3}
\cmidrule(lr){4-5}
Recognizer
& \shortstack{FAR =$10^{-3}$}
& \shortstack{FAR =$10^{-4}$}
& \shortstack{FAR =$10^{-3}$}
& \shortstack{FAR =$10^{-4}$} \\
\midrule
ArcFace
& 0.21588 & 0.30294 & 0.19628 & 0.25388 \\
AdaFace
& 0.24480 & 0.35714 & 0.19770 & 0.25562 \\
StyleID
& 0.37630 & 0.47013 & 0.45373 & 0.55375 \\
ArcFace-FT
& 0.21531 & 0.27064 & 0.31407 & 0.40871 \\
MagFace
& 0.31442 & 0.43047 & 0.26256 & 0.32810 \\
\bottomrule
\end{tabular}
\end{table}

\begin{table}[b]
\centering
\caption{
Ablation of the training objectives in AIFR.
}
\label{tab:objective_ablation}
\footnotesize
\setlength{\tabcolsep}{4.0pt}
\renewcommand{\arraystretch}{1.2}
\begin{tabular}{ccc|ccc}
\toprule
$\mathcal{L}_{\mathrm{id}}$
& $\mathcal{L}_{\mathrm{str}}$
& $\mathcal{L}_{\mathrm{pre}}$
& Rank-1 $\uparrow$
& TAR@$10^{-3}$ $\uparrow$
& TAR@$10^{-4}$ $\uparrow$ \\
\midrule

\checkmark & &
& 86.85 & 76.28 & 62.68 \\

& \checkmark &
& 84.93 & 69.18 & 53.98 \\

& & \checkmark
& 84.93 & 69.17 & 53.98 \\

\checkmark & & \checkmark
& 86.93 & 76.45 & 62.99 \\

\rowcolor{gray!10}
\checkmark & \checkmark & \checkmark
& \textbf{87.29}
& \textbf{76.82}
& \textbf{63.34} \\

\bottomrule
\end{tabular}
\end{table}

\begin{table*}[t]
\centering
\caption{
Ablation of the recovery designs in AIFR.
}
\label{tab:design_ablation}
\footnotesize
\setlength{\tabcolsep}{4.0pt}
\renewcommand{\arraystretch}{1.10}
\begin{tabular}{ccc|ccc}
\toprule
Fused-ID
& Adaptive strength
& Token Modulation
& Rank-1 $\uparrow$
& TAR@$10^{-3}$ $\uparrow$
& TAR@$10^{-4}$ $\uparrow$ \\
\midrule

\checkmark & &
& 86.77 & 76.47 & 63.30 \\

& \checkmark &
& 87.00 & 76.65 & 63.17 \\

& & \checkmark
& 86.94 & 76.44 & 62.96 \\

& \checkmark & \checkmark
& 87.20 & 76.78 & 63.29 \\

\rowcolor{gray!10}
\checkmark & \checkmark & \checkmark
& \textbf{87.29}
& \textbf{76.82}
& \textbf{63.34} \\

\bottomrule
\end{tabular}
\end{table*}

\begin{table}[t]
\centering
\caption{
Sensitivity to the loss weights.
The other hyperparameters are fixed when varying each weight.
}
\label{tab:loss_weight_ablation}
\footnotesize
\setlength{\tabcolsep}{4.5pt}
\renewcommand{\arraystretch}{1.08}
\begin{tabular}{ccccc}
\toprule
Parameter
& Value
& Rank-1 $\uparrow$
& TAR@$10^{-3}$ $\uparrow$
& TAR@$10^{-4}$ $\uparrow$ \\
\midrule

\multirow{3}{*}{$\lambda_{\mathrm{str}}$}
& 0.25
& 87.05 & 76.39 & 62.87 \\
& \textbf{0.5}
& \textbf{87.29}
& \textbf{76.82}
& \textbf{63.34} \\
& 1.0
& 87.07 & 75.95 & 62.10 \\

\midrule

\multirow{3}{*}{$\lambda_{\mathrm{pre}}$}
& \textbf{0.25}
& \textbf{87.29}
& \textbf{76.82}
& \textbf{63.34} \\
& 0.5
& 87.24 & 76.60 & 63.14 \\
& 1.0
& 87.10 & 76.15 & 62.85 \\

\bottomrule
\end{tabular}
\end{table}

\begin{table}[t]
\centering
\caption{
Sensitivity to the fusion coefficient $\alpha$ in the fused identity target.
}
\label{tab:alpha_ablation}
\footnotesize
\setlength{\tabcolsep}{6pt}
\renewcommand{\arraystretch}{1.08}
\begin{tabular}{cccc}
\toprule
$\alpha$
& Rank-1 $\uparrow$
& TAR@$10^{-3}$ $\uparrow$
& TAR@$10^{-4}$ $\uparrow$ \\
\midrule

0.0
& 86.90 & 76.55 & 63.20 \\

0.1
& 86.95 & 76.69 & 63.30 \\

\rowcolor{gray!10}
0.3
& \textbf{87.29}
& \textbf{76.82}
& \textbf{63.34} \\

0.5
& 87.14 & 76.64 & 63.21 \\

0.7
& 86.79 & 76.18 & 62.52 \\

0.9
& 87.11 & 76.09 & 62.42 \\

1.0
& 87.05 & 76.06 & 62.45 \\

\bottomrule
\end{tabular}
\end{table}

\begin{table*}[t]
\centering
\caption{Computational costs and gallery compatibility. Trainable parameters refer to the parameters optimized under
each method's training configuration. Training time reports additional training performed in this
study. -- denotes no additional training in this study (i.e., apply the pretrained backbone).}
\label{tab:efficiency}
\small
\setlength{\tabcolsep}{6pt}
\renewcommand{\arraystretch}{1.12}
\begin{tabular}{@{}lcccc@{}}
\toprule
Method &
\shortstack{Trainable\\params (M)} &
\shortstack{Training time\\(min)} &
\shortstack{Inference\\(ms/image)} &
\shortstack{Gallery\\re-encoding} \\
\midrule
ArcFace & 65.13 & -- & $7.05\pm0.13$ & No \\
AdaFace & 65.15 & -- & $9.45\pm0.34$ & Yes \\
ArcFace-FT & 65.13 & 21.31 & $7.03\pm0.11$ & Yes \\
StyleID & 4.70 & -- & $8.84\pm0.79$ & Yes \\
DiffClean+ArcFace & 93.56 &-- & $825.46\pm1.95$ & No \\
SAMC+ArcFace & 111.59 & 50.50 & $8.83\pm0.13$ & No \\
MAD+ArcFace & 182.51 & -- & $116536.25\pm210.91$ & No \\
\midrule
\textbf{AIFR} & \textbf{1.22} & 6.66 & $9.65\pm0.21$ & No \\
\bottomrule
\end{tabular}
\vspace{3pt}
\begin{minipage}{0.98\textwidth}
\footnotesize
ArcFace-FT and
AIFR average Makeup and Anime, while SAMC reports Makeup only.
\end{minipage}
\end{table*}

\begin{figure}[t]
    \centering
    \includegraphics[width=1\linewidth]{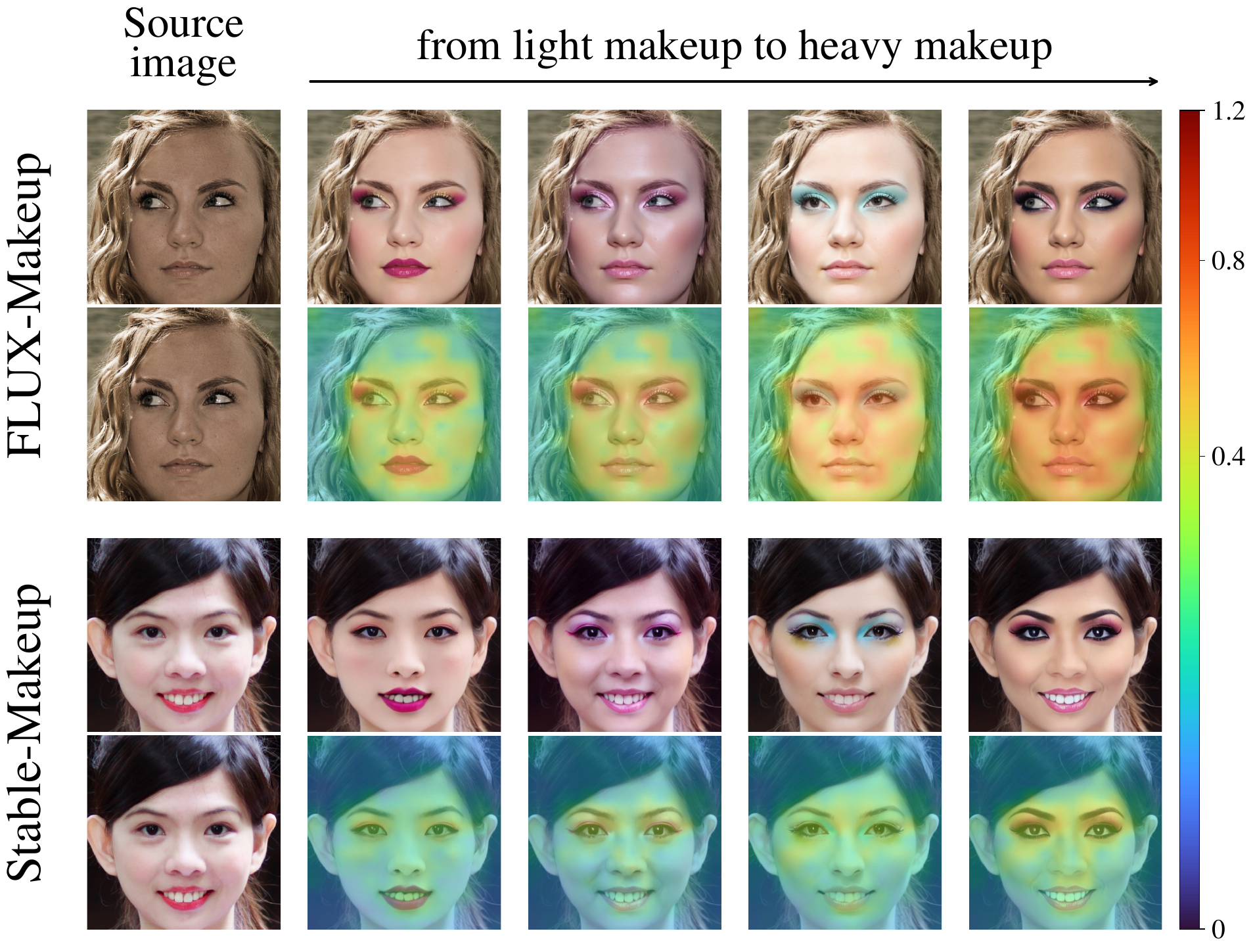}
    \caption{Visualization of AIFR on Makeup portraits across two personalization methods and five makeup types. For each method, the upper row shows stylized portraits, and the lower row shows the spatial distribution of AIFR feature corrections.}
    \label{fig:makeup_corrections}
\end{figure}

\begin{figure}[t]
    \centering
    \includegraphics[width=1\linewidth]{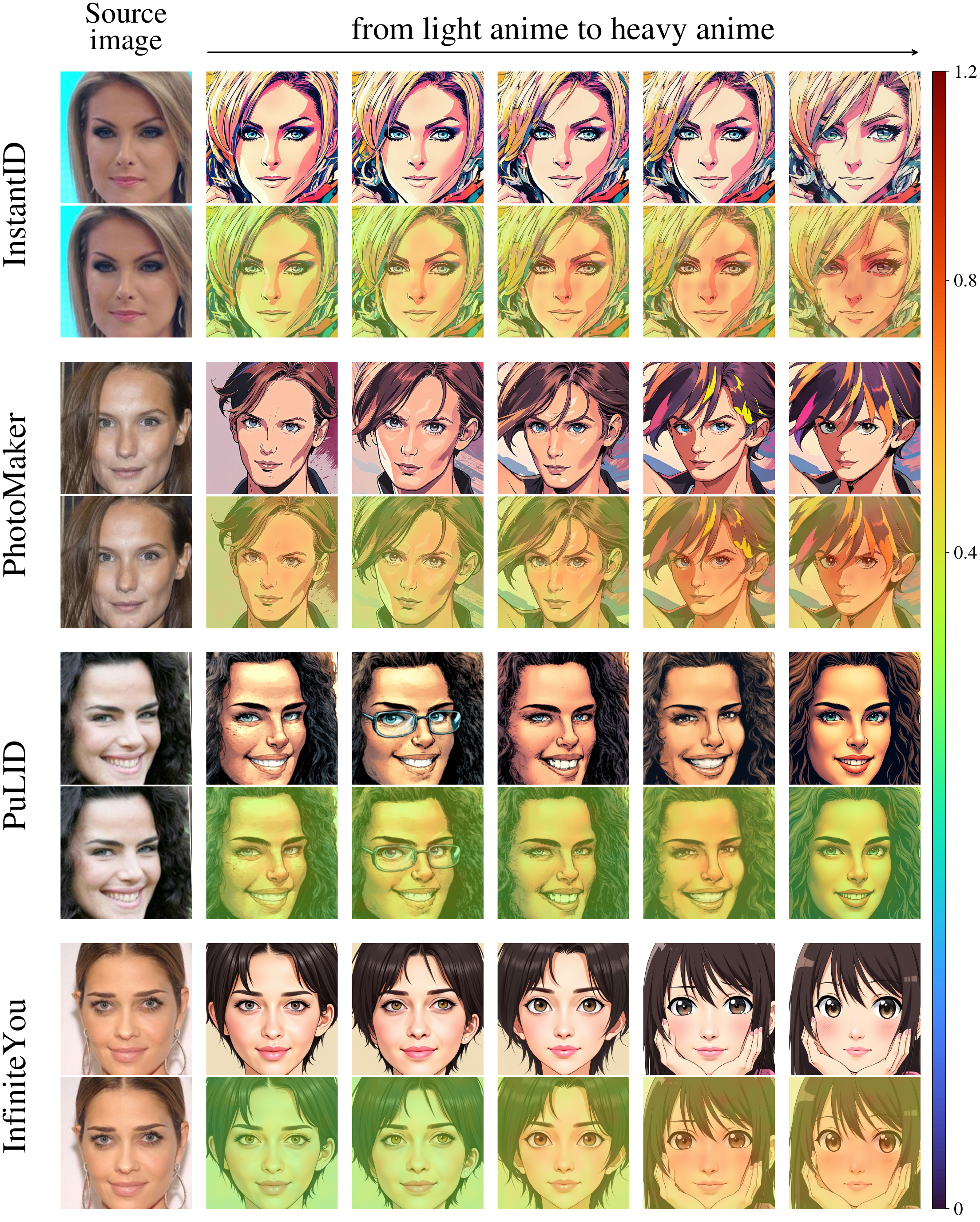}
    \caption{Visualization of AIFR on Anime portraits across four personalization methods and five stylization strengths. For each method, the upper row shows stylized portraits, and the lower row shows the spatial distribution of AIFR feature corrections.}
    \label{fig:anime_corrections}
\end{figure}

\begin{figure}[t]
    \centering
    \includegraphics[width=1\linewidth]{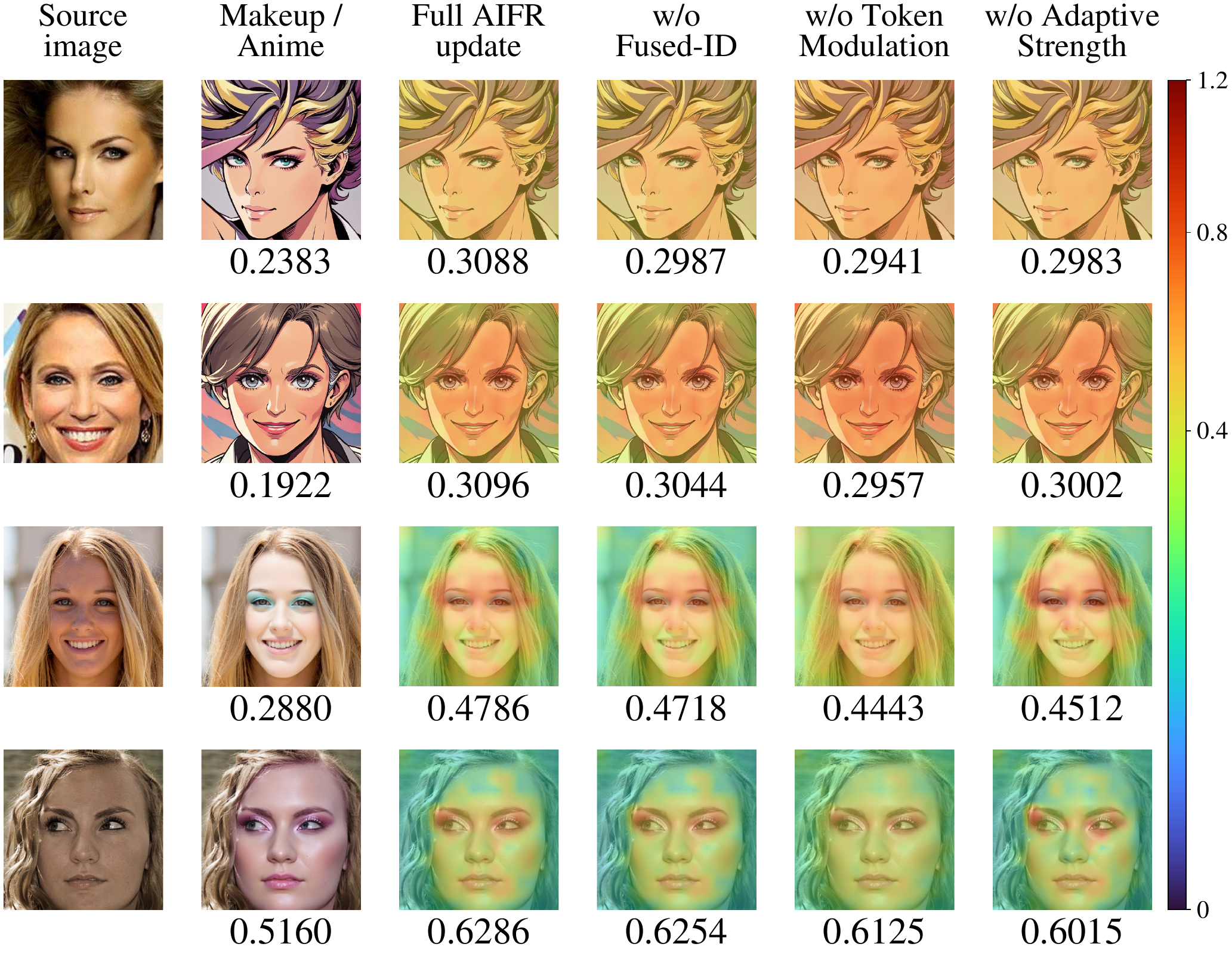}
    \caption{Qualitative ablation of AIFR components for full AIFR and variants. Heatmaps visualize the normalized spatial distribution of feature-update energy under a shared color scale. Values below the images report the cosine similarity between the recovered embedding and the clean reference embedding computed by ArcFace. Higher values indicate better identity preservation.}
    \label{fig:spatial_ablation}
\end{figure}

\section{Ablation Studies and Analysis}
\label{app:ablation}


This section provides additional ablations of the training
objectives, recovery designs, and hyperparameters, together
with qualitative analyses of feature corrections.
All ablations use the same evaluation sets of diffusion-based
makeup and anime portraits, covering two makeup generators
and four anime stylization generators across different
makeup types and stylization strengths.

\subsection{Training Objectives}

Table~\ref{tab:objective_ablation} shows that $\mathcal{L}_{\mathrm{id}}$ provides the strongest single-objective
result, while the complete objective achieves the best performance. The three losses have complementary roles: $\mathcal{L}_{\mathrm{id}}$ aligns recovered embeddings with clean
identity targets, $\mathcal{L}_{\mathrm{str}}$ guides sample-level correction strength, and $\mathcal{L}_{\mathrm{pre}}$ discourages
unnecessary changes to already reliable representations. Their combination supports identity recovery while limiting
the extent of correction.

\subsection{Recovery Designs}
Table~\ref{tab:design_ablation} compares the recovery components.
Without Fused-ID, the clean reference and identity prototype
provide separately weighted identity losses.
Disabling adaptive strength or token modulation sets
$s_i=1$ or $M_i=\mathbf{1}$, respectively.
Combining adaptive strength and token modulation improves
all three metrics over using either alone.
Adding Fused-ID provides a small further gain, with the
full design achieving the best results.
These findings support adapting both the overall correction
strength and its distribution across facial regions.

\subsection{Loss Weights}
Table~\ref{tab:loss_weight_ablation} examines sensitivity to the loss weights, showing that \(\lambda_{\mathrm{str}}=0.5\) and \(\lambda_{\mathrm{pre}}=0.25\) perform best among the tested settings. Increasing either weight to 1.0 lowers TAR at the stricter operating point, suggesting that excessive strength supervision or preservation constraints can compete with identity alignment.

\subsection{Fusion Weight}
Table~\ref{tab:alpha_ablation} examines the fusion weight in the Fused-ID design. The results show that \(\alpha=0.3\) achieves the best scores, with modest gains over the prototype-only supervision (i.e., $\alpha=0$). Verification performance declines at larger reference weights (\(\alpha\geq0.7\)), supporting a prototype-dominant target supplemented by the clean reference. 

\subsection{Qualitative Analysis}
 Fig.~\ref{fig:makeup_corrections} and~\ref{fig:anime_corrections} visualize spatial feature corrections across generators and appearance conditions. The heatmaps show relative RMS feature updates under a shared color scale. The visualizations show corrections for makeup concentrated around the eyes and mouth, with more extensive corrections in heavy makeup examples (Fig. \ref{fig:makeup_corrections}), whereas anime examples generally involve broader facial regions (Fig.~\ref{fig:anime_corrections}).
These patterns illustrate how spatially adaptive feature
correction accommodates different appearance conditions
by adjusting the distribution of updates across facial regions.

Fig.~\ref{fig:spatial_ablation} further compares correction
maps and ArcFace matching scores across component variants,
with the full model obtaining higher matching scores
in the displayed examples.





\section{Efficiency and Deployment Compatibility}
\label{app:efficiency}
Table~\ref{tab:efficiency} compares computational
costs and gallery compatibility on an NVIDIA
H100 NVL with batch size one and FP16 autocast.
AIFR introduces only 1.22M trainable parameters
while keeping the ArcFace backbone fixed,
allowing direct reuse of the existing gallery
embeddings without re-encoding.
The intermediate recovery module adds
2.60 ms/image to the original ArcFace inference
time (7.05 ms/image), resulting in
9.65 ms/image for AIFR.

\section{Discussion}
\label{app:discussion}
AIFR provides a lightweight approach to improving
face recognition under appearance alterations
without modifying the pretrained FR backbone.
As a plug-in module, it can be integrated into
existing FR pipelines, allowing service providers
to improve the recognition of altered faces
while directly reusing enrolled gallery embeddings.
This avoids the need for full-model retraining
or gallery re-encoding, reducing deployment costs.
Moreover, the recovered embeddings can also support
downstream embedding-conditioned tasks, such as face
reconstruction and personalization, further extending the utility of AIFR.

Although our experiments focus on diffusion-based
makeup and anime portraits, the proposed framework
is not inherently limited to these settings.
With appropriate training data, AIFR could
potentially be extended to other types of
appearance-altered faces.
Such extensions would require adapting the
recovery module to the target recognizer and
appearance variations.
Nevertheless, recovery performance ultimately
depends on the identity information preserved
in the altered inputs, and its effectiveness
under other appearance changes remains to
be investigated.

\section*{Ethical Considerations}
This work aims to improve face recognition under appearance alterations. However, the recovered identity features could also be misused for unauthorized identification or tracking, raising privacy and consent concerns. We encourage responsible use in authorized settings with appropriate privacy safeguards.


\end{document}